\documentclass[sigconf]{acmart}

\usepackage{booktabs}
\usepackage{multirow} 

\usepackage{makecell}
\usepackage{soul}
\usepackage{stfloats}
\usepackage{color}
\usepackage{xcolor}
\definecolor{darkgreen}{RGB}{83,129,53}
\definecolor{darkred}{RGB}{163,21,21}

\usepackage{tikz}
\usepackage{subfig}
\usepackage{tcolorbox}
\usepackage{pgfplots}
\usepackage{wrapfig}
\usepackage{mathrsfs}
\usepackage{array}
\usepackage{pgfplotstable}
\usepackage{pgf}
\usepackage[normalem]{ulem}
\usepackage{colortbl}
\usepackage{ulem}
\usepackage{xspace}

\usepackage{listings}

\usepackage{scalerel} 

\usepackage{latexsym}
\usepackage{graphicx}
\usepackage{amsmath}
\usepackage{multirow}
\usepackage{amsthm}
\usepackage{amsfonts}
\usepackage{bm}
\usepackage{booktabs}
\usepackage{algorithm}
\usepackage{algpseudocode}
\usepackage{verbatim}
\usepackage{tabularx}

\usepackage{latexsym}
\tcbuselibrary{breakable} 
\usepackage{fancyvrb}    
\tcbuselibrary{breakable,skins}   
\tcbuselibrary{listings}          
\usepackage{enumitem}  
\usepackage{hyperref}  
\usepackage{fvextra}   

\DefineVerbatimEnvironment{Verbatim}{Verbatim}{breaklines=true, breakanywhere=true}

\usepackage[utf8]{inputenc}

\usepackage{microtype}

\usepackage{graphicx}

\usepackage{multirow}
\usepackage{booktabs}
\usepackage{adjustbox}
\usepackage{colortbl}
\usepackage{xcolor}
\usepackage{booktabs}   
\usepackage{threeparttable} 
\usepackage{graphicx}   
\usepackage{xcolor}     

\definecolor{ModelGray}{HTML}{F2F2F2}
\definecolor{GeneralBlue}{HTML}{EDF3FF}
\definecolor{KnowledgeGreen}{HTML}{EEF7F0}

\usepackage{microtype}
\usepackage{hyperref}
\usepackage{adjustbox}
\usepackage{color}
\usepackage{xcolor}
\usepackage{tcolorbox}
\usepackage{colortbl}
\usepackage{multicol}
\usepackage{url}
\usepackage{booktabs}
\usepackage{amsmath}
\usepackage{enumitem}
\usepackage{graphicx}
\usepackage{lineno}
\usepackage{xspace}
\usepackage{algorithm}
\usepackage{algpseudocode}
\usepackage{array}
\usepackage{booktabs}
\usepackage{longtable}
\usepackage{arydshln}
\usepackage{caption}

\AtBeginDocument{%
  }

\setcopyright{acmlicensed}
\copyrightyear{2018}
\acmYear{2018}
\acmDOI{XXXXXXX.XXXXXXX}
\acmConference[Conference acronym 'XX]{Make sure to enter the correct
  conference title from your rights confirmation email}{June 03--05,
  2018}{Woodstock, NY}
\acmISBN{978-1-4503-XXXX-X/2018/06}

\begin{document}


\title{Towards Knowledgeable Deep Research: Framework and Benchmark}
\author{Wenxuan Liu}
\authornote{Contributed equally to this research.}
\additionalaffiliation{\institution{University of Chinese Academy of Sciences}}
\author{Zixuan Li}
\authornotemark[1]
\author{Long Bai}
\authornotemark[1]
\additionalaffiliation{\institution{National University of Singapore}}
\affiliation{\institution{State Key Laboratory of AI Safety, Institute of Computing Technology, Chinese Academy of Sciences}\city{Beijing}\country{China}}
\email{liuwenxuan2024z@ict.ac.cn}
\email{lizixuan@ict.ac.cn}
\email{bailong@ict.ac.cn}

\author{Chunmao Zhang}
\authornotemark[2]
\author{Fenghui Zhang}
\authornotemark[2]
\author{Zhuo Chen}
\authornotemark[2]
\author{Wei Li}
\affiliation{\institution{State Key Laboratory of AI Safety, Institute of Computing Technology, Chinese Academy of Sciences}\city{Beijing}\country{China}}

\author{Yuxin Zuo}
\authornotemark[2]
\author{Fei Wang}
\author{Bingbing Xu}
\author{Xuhui Jiang}
\affiliation{\institution{State Key Laboratory of AI Safety, Institute of Computing Technology, Chinese Academy of Sciences}\city{Beijing}\country{China}}

\author{Jin Zhang}
\authornotemark[2]
\author{Xiaolong Jin}
\authornotemark[2]
\authornote{Xiaolong Jin is the corresponding author.}
\author{Jiafeng Guo}
\authornotemark[2]
\affiliation{\institution{State Key Laboratory of AI Safety, Institute of Computing Technology, Chinese Academy of Sciences}\city{Beijing}\country{China}}
\email{jinxiaolong@ict.ac.cn}

\author{Tat-Seng Chua}
\affiliation{\institution{National University of Singapore}\city{Singapore}\country{Singapore}}

\author{Xueqi Cheng}
\authornotemark[2]
\affiliation{\institution{State Key Laboratory of AI Safety, Institute of Computing Technology, Chinese Academy of Sciences}\city{Beijing}\country{China}}

\renewcommand{\shortauthors}{Wenxuan Liu et al.}

\begin{abstract}
Deep Research (DR) requires LLM agents to autonomously perform multi-step information seeking, processing, and reasoning to generate comprehensive reports.
 In contrast to existing studies that mainly focus on unstructured web content, a more challenging DR task should additionally utilize structured knowledge to provide a solid data foundation, facilitate quantitative computation, and lead to in-depth analyses.
In this paper, we refer to this novel task as Knowledgeable Deep Research (KDR), which requires DR agents to generate reports with both structured and unstructured knowledge.
Furthermore, we propose the Hybrid Knowledge Analysis framework (HKA), a multi-agent architecture that reasons over both kinds of knowledge and integrates the texts, figures, and tables into coherent multimodal reports.
The key design is the Structured Knowledge Analyzer, which utilizes both coding and vision-language models to produce figures, tables, and corresponding insights.
To support systematic evaluation, we construct KDR-Bench, which covers 9 domains, includes 41 expert-level questions, and incorporates a large number of structured knowledge resources (e.g., 1,252 tables). We further annotate the main conclusions and key points for each question and propose three categories of evaluation metrics including general-purpose, knowledge-centric, and vision-enhanced ones.
Experimental results demonstrate that HKA consistently outperforms most existing DR agents on general-purpose and knowledge-centric metrics, and even surpasses the Gemini DR agent on vision-enhanced metrics, highlighting its effectiveness in deep, structure-aware knowledge analysis.
Finally, we hope this work can serve as a new foundation for structured knowledge analysis in DR agents and facilitate future multimodal DR studies. 

\end{abstract}

\ccsdesc[500]{Information systems~Information retrieval}

\keywords{Large Language Models, Deep Research}

\maketitle

\section{Introduction}

Recently, Large Language Model (LLM) agents have demonstrated remarkable capabilities across a variety of complex tasks, including mathematics~\cite{liu2025mmagentllmagentsrealworld}, software engineering~\cite{bouzenia2025understandingsoftwareengineeringagents}, and scientific discovery~\cite{huang2025cascadecumulativeagenticskill}. Among these tasks, the Deep Research (DR) task has emerged as a critical area of focus due to its ability to facilitate sophisticated, high-stakes applications. Compared to traditional information retrieval, the DR task requires LLM agents to autonomously perform multi-step information seeking, processing, and reasoning to generate comprehensive, evidence-grounded reports.

However, existing DR agents either overly depend on web search \cite{li2025webthinker} or adopt predefined tools to generate short-form responses~\cite{tongyidr}, both of which fall short in flexibly reasoning over large-scale structured knowledge (e.g. table and graphs) and make it challenging to conduct comprehensive research for answering questions such as ``What factors account for the regional differences in the investment of ESG worldwide in 2025?''. For such questions, structured knowledge is essential because it can provide a solid data foundation, facilitate quantitative computation, and lead to in-depth analyses. We refer to this challenging DR task as Knowledgeable Deep Research (KDR).

To facilitate this task, we propose the Hybrid Knowledge Analysis framework (HKA), a multi-agent architecture that is able to reason over both structured and unstructured knowledge. Specifically, it consists of four LLM-based sub-agents, namely, Planner, Unstructured Knowledge Analyzer, Structured Knowledge Analyzer, and Writer. 
Given a research question, the Planner first decomposes it into several subtasks and controls the workflow. For each subtask, the Planner iteratively generates tool calls to invoke either the Unstructured Knowledge Analyzer or the Structured Knowledge Analyzer. Subsequently, the Unstructured Knowledge Analyzer and the Structured Knowledge Analyzer generate supporting materials, including text, figures, and tables, via leveraging corresponding knowledge sources. Finally, the Writer aggregates the multimodal materials from all subtasks, resolves the conflicts among them, and produces a coherent and comprehensive report. In this framework, our key design is the Structured Knowledge Analyzer, which adopts a code model to generate code for producing multimodal materials and a vision-language model to generate corresponding insights.

To support systematic evaluation, we construct KDR-Bench, a comprehensive benchmark that covers 9 domains, including Agriculture, Politics \& Economics, Energy \& Environment, Finance \& Insurance, Metals \& Electronics, Society, Art, Technology, and Transportation. Using a human-in-the-loop strategy, we curate 41 expert-level research questions and aggregate a structured knowledge base of 1,252 tables. Furthermore, we annotate the main conclusions and key points for each question.
Based on these data and annotations, we develop three categories of evaluation metrics, including general-purpose, knowledge-centric, and vision-enhanced ones, which measure the ability of DR agents to utilize both unstructured and structured knowledge in an LLM-as-a-Judge setting.

Finally, we evaluate 12 different baselines and HKA on the KDR-Bench.
The baselines include LLMs with search tools, closed-source DR agents, and open-source DR agents.
The experimental results show that HKA outperforms most baselines, including product-level DR agents.
Since HKA is able to generate multimodal content, we further evaluate the generated reports using a multimodal large language model (MLLM) as the judge.
The results show that HKA even outperforms the state-of-the-art DR agent, further validating its effectiveness.
These findings also reveal the limitations of traditional evaluation methods when applied to multimodal reports.
We hope this work can serve as a new foundation for structured knowledge analysis in DR agents and facilitate future multimodal DR studies.
The main contributions of this work are summarized as follows:
\begin{itemize}
    \item We introduce the KDR task, which challenges DR agents to reason over structured and unstructured knowledge and produce thorough research reports.
    \item We propose HKA that integrates both structured and unstructured knowledge via a multi-agent framework to generate comprehensive multimodal reports. With a Structured Knowledge Analyzer that is based on both coding and vision-language models, HKA can produce figures, tables, and corresponding insights beyond text outputs. 
    \item We construct KDR-Bench, an expert-level benchmark across 9 domains, including 41 questions with extensive structured knowledge, along with an LLM-based evaluation framework for assessing knowledge utilization in each report.
    \item Experimental results demonstrate that HKA outperforms most DR agents on general-purpose and knowledge-centric metrics, and even surpasses Gemini on vision-enhanced metrics, highlighting its effectiveness in multimodal report generation.
\end{itemize}

\section{Related Work}

\paragraph{Deep Research Agent}
Deep Research agents aim to support in-depth investigation of user queries through large-scale information retrieval, organization, and long-form writing~\cite{shi2025deepresearchsystematicsurvey,xu2025comprehensivesurveydeepresearch}. 
This paradigm has gained significant traction in industrial applications, such as Gemini~\cite{google_gemini_deep_research_doc_2025} and Perplexity~\cite{perplexity_deep_research_blog_2025}, where these capabilities are regarded as a hallmark of advanced agentic reasoning and tool proficiency~\cite{zhang2025landscapeagenticreinforcementlearning}. 
In parallel, the open-source community strives to narrow the gap with proprietary models. Existing efforts generally fall into two categories: constructing robust multi-agent workflows to emulate closed-source systems~\cite{langchain_open_deep_research_2025,jin2025finsightrealworldfinancialdeep,chen2025BrowseCompPlus}, or employing agentic reinforcement learning~\cite{wu2025agenticreasoningstreamlinedframework,dong2025agenticentropybalancedpolicyoptimization,yao2026oresearcheropenendeddeep} to train LLMs to master complex tool usage like information seeking and long-form writing~\cite{zheng2025deepresearcher,li2025webthinker,tongyidr, li2025websailor}. Nevertheless, most existing deep research agents primarily operate over unstructured web resources, with limited support for computation and reasoning over structured knowledge.

\paragraph{Deep Research Benchmarks}
Deep Research benchmarks are generally categorized into two types: complex problem-solving and long-form report generation. Representative benchmarks for problem-solving include Humanity's Last Exam (HLE)~\cite{HLE} and BrowserComp~\cite{wei2025browsecompsimplechallengingbenchmark,zhou2025browsecompzhbenchmarkingwebbrowsing,chen2025BrowseCompPlus}, which primarily evaluate capabilities in multi-step reasoning and information seeking. For report generation tasks, prominent benchmarks include DeepResearch Bench~\cite{du2025deepresearchbenchcomprehensivebenchmark} and Personal DR~\cite{liang2025personalizeddeepresearchbenchmarks}.
However, the questions in the report-oriented benchmarks typically prioritize textual information aggregation, which falls short of providing a fine-grained evaluation of an agent's ability to use knowledge for quantitative analysis and the derivation of novel conclusions. KDR-Bench is designed to evaluate the capability of knowledge analysis in DR agents.

\section{Hybrid Knowledge Analysis Framework}
\label{sec:framework}

\begin{figure*}
    \centering
    \includegraphics[width=\textwidth]{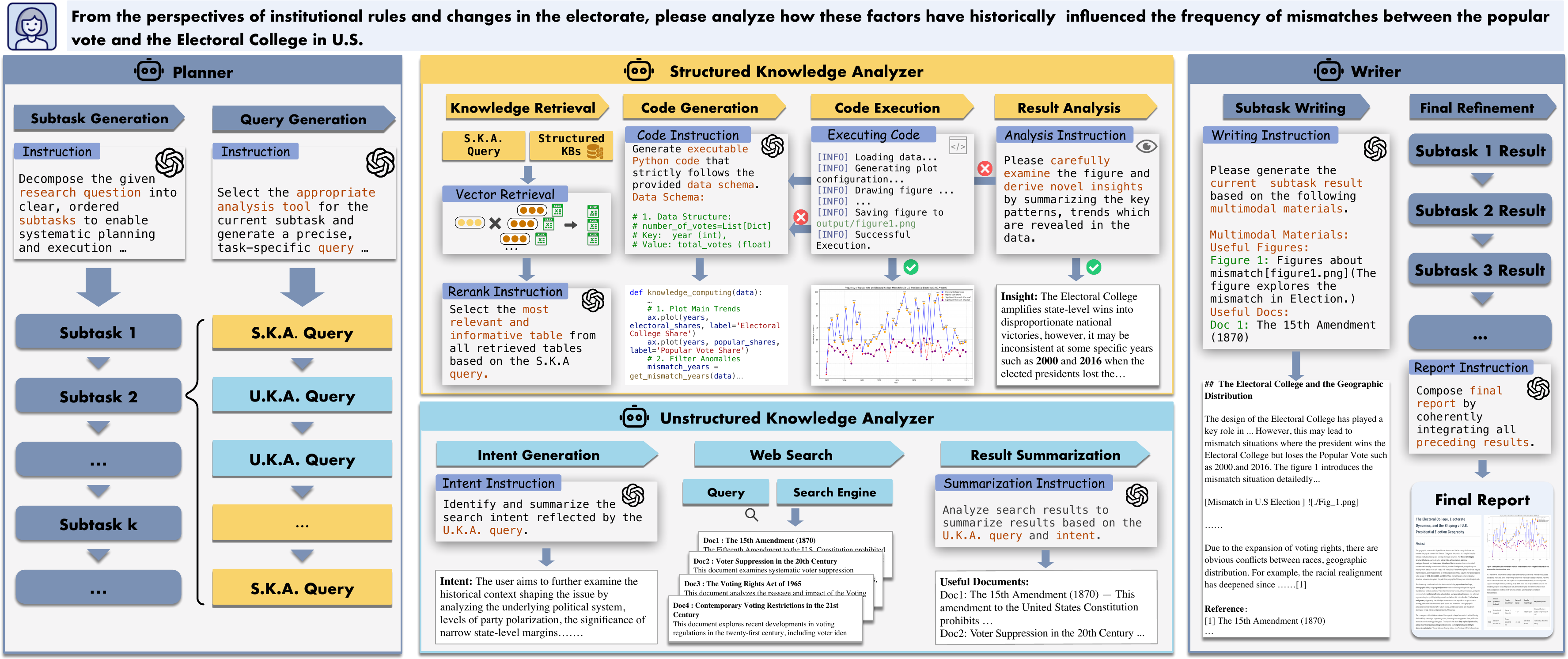}
    \caption{Demonstration of the proposed HKA framework. }
    \label{fig:hka}
\end{figure*}

\subsection{Problem Formulation}
Given a research question $q$, the Knowledgeable Deep Research (KDR) task requires an LLM agent to generate a long-form, multimodal output $y$ (i.e., a report) based on multi-step reasoning over both structured knowledge sources $\mathcal{S}$ and unstructured knowledge sources $\mathcal{U}$.
The total reasoning trajectory is denoted as $\mathcal{R}=(r_1, ..., r_i, ..., r_T)$. Each reasoning step $r_i$ autonomously invokes tools from an available tool set $\mathcal{T}$ related to the above two kinds of knowledge sources. The overall process is formalized by the following conditional probability,
\begin{equation}
P(y \mid q, \mathcal{S}, \mathcal{U})= \prod_{i=1}^{T} P(r_i \mid r_{<i}, q, \mathcal{T}, \{\mathcal{O}_{\tau}\}_{\tau<i}),
\end{equation}
where $\{\mathcal{O}_{\tau}\}_{\tau<i}$ denotes the outputs of all tool calls before step $i$. $r_{<i}$ represents the reasoning steps generated before position $i$. Unlike traditional DR tasks that primarily focus on unstructured knowledge sources (e.g., web pages), the KDR task requires agents to additionally utilize structured knowledge sources $\mathcal{S}$, such as tables, beyond web pages.

\subsection{Overview of HKA}
Existing deep research frameworks~\cite{li2025webthinker, prabhakar2025enterprisedeepresearch} typically do not distinguish between different types of knowledge and can only perform shallow analysis over structured knowledge sources.
Therefore, they lack the capability to efficiently process large-scale structured data, perform complex computation, and further obtain novel insights effectively.
To address these limitations, we propose HKA, a multi-agent framework that leverages structured knowledge to provide a solid data foundation, facilitate quantitative computation, and enable in-depth analyses.
In addition, HKA integrates unstructured knowledge sources to produce comprehensive research reports.
As shown in Figure~\ref{fig:hka}, HKA consists of four sub-agents: 1) the Planner, which designs subtasks, invokes the following three sub-agents, and controls the workflow; 2) the Structured Knowledge Analyzer, which computes over structured knowledge via a code LLM and a VLM; 3) the Unstructured Knowledge Analyzer, which retrieves and summarizes unstructured knowledge sources; and 4) the Writer, which integrates the supporting materials from the two analyzers into subtask results and composes the final report.

\subsection{Planner}
Given a research question $q$, the Planner automatically performs task planning and dynamically invokes other agents to collaboratively complete the report generation process.
Specifically, the Planner first decomposes the question into a sequence of fine-grained subtasks.
Then, for each subtask, the Planner autonomously decides which type of knowledge source is currently required, and generates a tool call to invoke the corresponding knowledge analyzer.
For the Unstructured Knowledge Analyzer (U.K.A.), the Planner generates a U.K.A. query as input and passes the historical state as context.
For the Structured Knowledge Analyzer (S.K.A.), the Planner generates an S.K.A. query and similarly provides the historical state.
After the Planner gathers the supporting materials via iterative tool calls, it invokes the Writer to organize these materials and write a coherent subtask result corresponding to that subtask.
Finally, after all subtasks are completed, the Planner invokes the Writer to refine the overall report based on all subtask results.

\subsection{Unstructured Knowledge Analyzer} 
The Unstructured Knowledge Analyzer is designed to search and integrate unstructured knowledge into supporting materials based on the U.K.A. query from the Planner.
Following~\citet{prabhakar2025enterprisedeepresearch}, in this paper, we primarily focus on web content retrieved via search engines. 
Specifically, the Unstructured Knowledge Analyzer consists of three main steps, namely, Intent Generation, Web Search, and Result Summarization.

\paragraph{Intent Generation}
Since the U.K.A. queries generated by the Planner are usually short and lack specific details, it is difficult to obtain precise information from massive web pages with only these queries.
Therefore, we prompt the LLM to expand each U.K.A. query into a detailed search intent based on the current subtask~\cite{jagerman2023queryexpansionpromptinglarge,li2025webthinker}, so that we can extract relevant information from web pages.

\paragraph{Web Search}
Based on the U.K.A. query and generated search intent, this sub-agent retrieves web pages through an existing search engine.
Then, the sub-agent converts the web pages into Markdown, an LLM-friendly format.

\paragraph{Result Summarization}
Finally, the sub-agent summarizes relevant information, including key data, descriptions, and conclusions, from the retrieved Markdown-formatted web pages according to the subtask, the U.K.A. query, and the search intent.
To save the context length for the Planner, only the summarized information is visible to the Planner, whereas the original web pages are chunked and stored for the Writer. 

\subsection{Structured Knowledge Analyzer}
Given the S.K.A. query and the historical state from the Planner, the Structured Knowledge Analyzer aims to retrieve structured knowledge relevant to the query, perform quantitative computation, and produce multimodal analysis results with corresponding insights.
Since the table is a typical form of structured knowledge, we use it as an example to describe how this sub-agent works.
This sub-agent consists of four main steps, namely, Knowledge Retrieval, Code Generation, Code Execution, and Result Analysis.

\paragraph{Knowledge Retrieval}
To obtain the structured knowledge relevant to the current subtask, the sub-agent first uses the S.K.A. query to retrieve tables based on their descriptions (including titles and summaries).
Specifically, we adopt a retrieve-and-rerank pipeline for this step, where a dense retriever recalls the top-$k$ tables according to the similarity scores between the query and the table descriptions.
Then, an LLM re-ranks the retrieved tables to select the most relevant table.
To avoid redundant analysis of the same table, we filter out previously used tables during retrieval.

\paragraph{Code Generation and Execution}
To conduct flexible computation over the retrieved table, we adopt a code LLM to generate and execute customized computation code for different questions and subtasks.
Instead of appending all data in the table to the prompt, we summarize the schema of each table in the form of comments and convert the table into a list of corresponding objects (e.g., ``real\_gdp\_growth\_of\_canada = [...]'').
In the Code Generation step, only the comments are given in the prompt, while the specific objects are invisible.
In the Code Execution step, the objects are injected before the computation code.
Through such a strategy, the code LLM can understand how to access the objects with much fewer tokens compared to directly injecting the entire table into the prompt.
In addition, we observe that the generated code sometimes fails to execute. Thus, we adopt a retry mechanism in the Code Execution step.
If the code fails to run successfully, the error messages will also be attached to regenerate the computation code, unless a predefined maximum number of retries is reached.
As a result, the execution failure rate decreases from 31.7\% to 0.51\%.

\paragraph{Result Analysis}
Since the Code Execution step may produce figures or tables, we adopt a vision-language model (VLM) to analyze these results and derive corresponding insights.
The Writer will use the figures and tables in the report, and generate textual analysis based on the insights.
In practice, we found that the Code Execution results are sometimes empty, or include visually incorrect or question-irrelevant outputs.
Therefore, we also adopt a retry mechanism in this step.
Specifically, the VLM first determines whether the sub-agent should regenerate the computation code using the code model. Once the generated figures pass VLM-based validation and the corresponding conclusions are produced, the validated materials are forwarded to the Planner.
As a result, the failure rate decreases from 55.5\% to 1.7\%.

\subsection{Writer}
The Writer aims to organize the supporting materials produced by the Unstructured and Structured Knowledge Analyzers into a coherent and comprehensive report.
Considering the limited context length, we divide the writing process into two steps, namely, Subtask Writing and Final Refinement.

\paragraph{Subtask Writing}
When each subtask is completed, the Writer immediately integrates the supporting materials for this subtask, resolves the conflicts among them, and writes a subtask result.
In practice, although we emphasize the importance of multimodal materials in the prompt, the Writer still tends to exclude them from the output, partly due to the dominance of textual information from web sources.
Therefore, we first generate an outline which retains most multimodal materials, and then fill in the textual content to produce the subtask result.

\paragraph{Final Refinement}
When all subtasks are completed, the Writer composes all the subtask results to form a complete and coherent report.
Considering that simply combining the results from predefined subtasks may not produce a high-quality report, the Writer is required to adjust the report structure based on these results, remove redundancies, resolve logical inconsistencies, and finally generate a well-structured, comprehensive, and coherent report.

\section{KDR-Bench}\label{sec:benchmark}

\begin{figure*}
    \centering
    \includegraphics[width=0.85\linewidth]{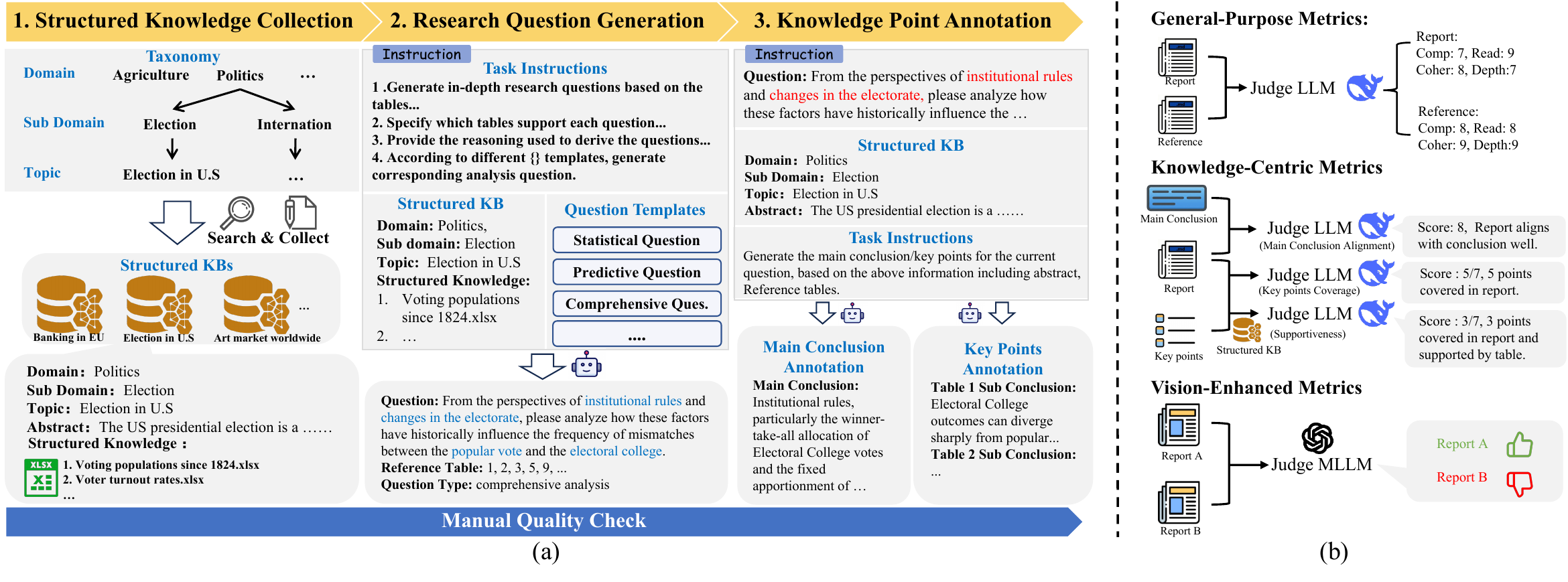}
    \caption{The construction procedure of KDR-Bench: (a) Dataset construction process; (b) Evaluation framework.}
    \label{fig:dataset_construction}
\end{figure*}

The proposed KDR-Bench includes two parts, a comprehensively annotated dataset and a corresponding evaluation framework.
To efficiently obtain an expert-level dataset, we introduce a human-in-the-loop data construction process, which incorporates an LLM with human review and revision.
To emphasize the knowledge utilization during evaluation, we introduce a knowledge-enhanced evaluation framework, which extends existing general-purpose metrics with our newly proposed knowledge-centric and vision-enhanced metrics.

\subsection{Dataset Construction}

As shown in Figure~\ref{fig:dataset_construction}, the dataset construction process consists of three main steps, namely, Data Collection, Question Generation, and Knowledge Point Annotation.
In what follows, we will introduce these three steps in more detail.

\subsubsection{Structured Knowledge Collection}

\begin{figure}
    \centering
    \includegraphics[width=0.95\linewidth]{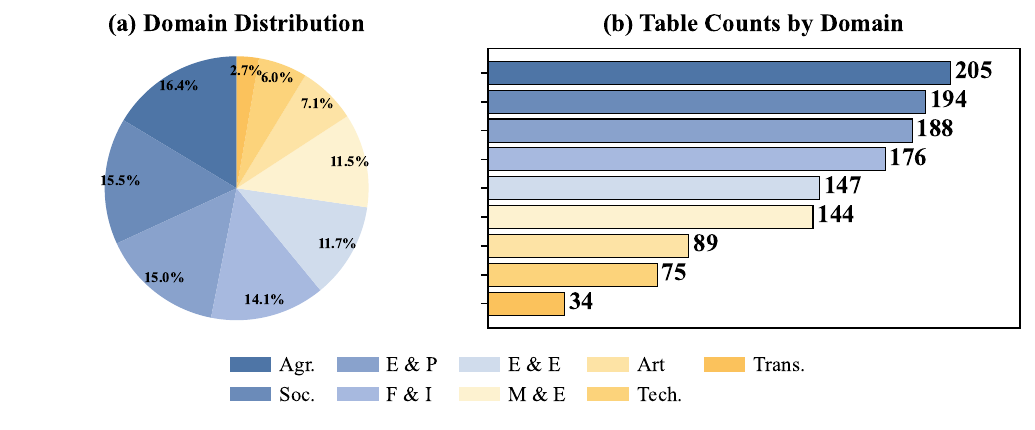}
    \caption{The statistics on tables in KDR-Bench.}
    \label{fig:table_statistics}
\end{figure}

We collect the structured knowledge from an online data analytics platform~\footnote{ https://www.statista.com. The website restricts commercial use of its data. All data collected in this work are publicly available. }.
As illustrated in Figure~\ref{fig:dataset_construction}, the data are organized via a three-level hierarchy, namely, domain, sub-domain, and topic.
Each topic is associated with an abstract which describes key concepts about the topic and a set of tables which summarize key indicators based on expert knowledge.
Since the table contents are not publicly accessible on the platform, we manually retrieve the corresponding data sources via search engines.
Further, we adopt a code model (Qwen3-Coder~\cite{qwen3}) to generate the processing code that converts table data into Python objects, such as dictionaries or lists, accompanied by a comment-style schema description specifying field names.
Finally, we obtain 9 domains, 18 sub-domains, 41 topics, accompanied by 1,252 tables.
The domains include Agriculture (Agr.), Politics \& Economics (P\&E), Energy \& Environment (E\&E), Finance \& Insurance (F\&I), Metals \& Electronics (M\&E), Society (Soc.), Art, Technology (Tech.), and Transportation (Trans.).
The number and distribution of tables in each domain are shown in Figure~\ref{fig:table_statistics}.

\subsubsection{Research Question Generation}
We create the research questions based on the collected data.
To avoid overly long contexts, we replace raw table contents in prompts with LLM-generated table summaries that describe key data characteristics and trends.
To promote diversity in question types, we generate candidate questions from six distinct templates, each targeting a different analytical behavior, including Comprehensive (holistic synthesis of theory, methods, and evidence), Predictive (forecasting future trends based on historical patterns), Categorization (systematic classification and typology construction), Statistical (quantitative relationship and pattern analysis), Attributive (causal mechanism and factor attribution analysis), and Comparative (comparison across entities, contexts, or time periods).
Then, a human expert reviews the candidate questions and selects the most suitable one as the final question.
Finally, we further refine the questions and obtain 41 high-quality questions in total.

In addition, we observe that the generated questions are often vague, leading to unclear analytical scopes.
To enhance the quality of the questions, we instruct the LLM with detailed guidelines, including specific temporal scopes (e.g., ``from 2010 to 2020''), geographic boundaries (e.g., ``the U.S. and China''), and analytical perspectives (e.g., ``from the perspectives of institutional rules and changes in the electorate'').

\subsubsection{Knowledge Point Annotation}
To evaluate the knowledge utilization of the DR agents, we further annotate each question with a main conclusion and a set of key points.
The main conclusion provides a high-level response to the question, which is generated based on the question, topic abstract, and corresponding table summaries.
Each key point describes the detailed analysis grounded in a specific table, which is generated based on the main conclusion and the corresponding table.
Both main conclusions and key points are reviewed by human experts in terms of professionalism, factuality, and analytical depth.
In total, we obtain 41 main conclusions and 261 key points.

\subsection{Evaluation Framework}
\label{sec:evaluation_framework}
Existing evaluation methods focus on the general-purpose metrics such as comprehensiveness or readability.
However, they fall short in justifying whether the DR agents utilize the appropriate knowledge and derive reasonable conclusions.
Although some studies~\cite{du2025deepresearchbenchcomprehensivebenchmark,wan2025deepresearcharenaexamllms} have investigated the trustworthiness of unstructured knowledge sources, the evaluation of structured knowledge utilization in DR remains largely unexplored.
To systematically evaluate the DR agents on the proposed dataset, we propose a knowledge-enhanced evaluation framework, which consists of three categories of metrics, namely, general-purpose metrics, knowledge-centric metrics, and vision-enhanced metrics.

\subsubsection{General-Purpose Metrics}
Following the existing studies~\cite{du2025deepresearchbenchcomprehensivebenchmark}, we adopt RACE to evaluate multiple dimensions of a generated report against a reference report.
Each dimension consists of a set of detailed evaluation criteria with different criterion weights.
The criterion-level score of a criterion $c$ is calculated via comparing the generated report $y_{gen}$ with the reference report $y_{ref}$ using an LLM as the judge:
\begin{equation}
    (s_{y_{gen},c}, s_{y_{ref},c}) = \mathbb{J}_G(y_{gen}, y_{ref}, c),
\end{equation}
where $\mathbb{J}_G(\cdot)$ is the judge function that assigns scores for the generated and reference reports, respectively, according to the criterion.

Then, the dimension score of dimension $d$ is the weighted sum of all its criteria scores.
Similarly, the report intermediate score $s_{y,Int}$ is the weighted sum of all its dimension scores.
Finally, the overall score of the generated report $s_{y_{gen}}$ is calculated as follows:
\begin{equation}
    s_{y_{gen}} = \frac{s_{y_{gen}, Int}}{s_{y_{gen}, Int} + s_{y_{ref}, Int}}.
\end{equation}
Following the convention, we report both the overall average score and the per-dimension scores.
In particular, we replace the original ``Instruction-Following'' dimension with ``Coherence'' (Coher.) to focus on the overall logical consistency of the reports, and then regenerate the criteria for this dimension.
We retain the other dimensions, namely, Comprehensiveness (Comp.), Depth, and Readability (Read.).
Since we have modified the dimensions, we assign equal weights to each dimension to avoid regenerating all weights.

\subsubsection{Knowledge-Centric Metrics}
\label{sec:knowledge_centric_metrics}


Based on the main conclusions and key points, we propose three metrics, namely, main conclusion alignment, key point coverage, and key point supportiveness.

Main conclusion alignment (Main.) measures how well the generated report aligns with the main conclusion.
We utilize an LLM as the judge to score the generated report on a 0-10 scale:
\begin{equation}
    s_{Main} = \mathbb{J}_K^{(1)}(y_{gen}, M, q),
\end{equation}
where $M$ denotes the main conclusion; $q$ is the question; $\mathbb{J}_K^M(\cdot)$ is the judge function.
The overall score is calculated by averaging the scores across all the reports.
To capture more fine-grained differences between agents, we multiply the overall score by 10 to rescale it to the range of 0-100.
Although the main conclusion is generated via the question, abstract, and corresponding tables, it is not strongly bound to the tables.
Therefore, an agent can still obtain a high score even if it only utilizes unstructured knowledge.

Key point coverage (Key.) measures how well the generated report covers each key point.
Given the key point list $P$, we utilize the LLM to judge whether each key point is covered in the report:
\begin{equation}
    s_{key} = \frac{\sum \mathbb{J}_K^{(2)}(y_{gen}, P, q)}{|P|},
\end{equation}
where $\mathbb{J}_K^{(2)}(\cdot)$ is the judge function returning a list of binary indicators, with 1 indicating that the key point is covered and 0 otherwise.
Since the key points are generated directly via the corresponding table, this metric is more related to structured knowledge utilization.
However, a DR agent can still cover some of the key points via utilizing the unstructured knowledge.

Based on the above metrics, we further introduce the key point supportiveness score (Support.), which considers not only the  coverage, but also whether the corresponding tables are retrieved and analyzed.
This metric measures whether the agent derives correct key points by leveraging the appropriate tables.
Formally, it is calculated as follows:
\begin{equation}
    s_{sup} = \frac{\sum \mathbb{J}_K^{(2)}(y_{gen}, P, q) \bigwedge \mathbb{J}_K^{(3)}(y_{gen}, T, q)}{|P|},
\end{equation}
where $T$ is the set of ground-truth tables associated with question $q$; $\mathbb{J}_K^{(3)}(\cdot)$ is the judge function that returns a list of binary indicators specifying whether the table corresponding to each key point is utilized in the report (1 for used and 0 otherwise); and $\bigwedge$ is the element-wise conjunction operation.
Since this metric requires retrieving the tables, it mainly reflects the ability of an agent to utilize structured knowledge.

\subsubsection{Vision-enhanced Metrics}
In existing approaches, the generated reports are represented in markdown format and evaluated via a text-based LLM.
Under this setting, the figures are represented as insertion markers.
Thus, the LLM fails to actually understand the content of the figures.
Therefore, we compile the reports into PDFs and utilize a multimodal LLM (MLLM) as the judge to take the visual features, such as report layout and figure content, into consideration.
Specifically, considering the cost and efficiency, we adopt pairwise comparison, where an agent pair $(A,B)$ is compared to calculate the win rate $S_V(A,B)$ using an MLLM judge as follows:
\begin{equation}
    S_V(A,B)=\frac{\sum_{i=1}^N \mathbb{J}_{V}(y_{i,A}, y_{i,B})}{N},
\end{equation}
where $N$ is the number of questions; $y_{i,X}$ is the report generated by agent $X\in\{A,B\}$ for the $i$-th question; the indicator function $\mathbb{J}_{V}(\cdot)$ outputs 1 if the judge prefers $y_{i,A}$ over $y_{i,B}$, and 0 otherwise.
We report both the overall win rate and per-domain win rates.

\begin{table*}[h]
    \centering
    \caption{Results on General-purpose and Knowledge-centric metrics. The best results are in boldface, and the second-best results are underlined.}
    \resizebox{0.65\linewidth}{!}{
    \begin{tabular}{lcccccccc}
        \toprule
        & \multicolumn{5}{c}{General-Purpose} & \multicolumn{3}{c}{Knowledge-Centric}
        \\
        \multirow{-2}{*}{Methods} & Avg. & Comp. & Depth & Coher. & Read. & Main. & Key. & Support.
        \\
        \midrule
        \multicolumn{9}{c}{\textit{LLM with search tools}}
        \\
        Hunyuan2.0 & 40.6 & 40.9 & 38.8 & 41.4 & 41.2 & 61.0 & 46.7 & -
        \\
        GLM4.6 & 40.8 & 40.8 & 36.4 & 42.6 & 43.5 & 59.1 & 40.1 & -
        \\
        Qwen3-Max & 44.6 & 43.9 & 40.5 & 46.6 & 47.4 & 64.2 & 43.5 & -
        \\
        Minimax-M2 & 44.7 & 44.8 & 41.1 & 46.8 & 46.0 & 58.2 & 50.5 & -
        \\
        \midrule
        \multicolumn{9}{c}{\textit{Closed-source DR Agents}}
        \\
        OpenAI & 41.3 & 41.4 & 38.0 & 43.1 & 42.7 & 63.5 & 44.1 & -
        \\
        Perplexity & 46.5 & 47.2 & 44.8 & 47.5 & 46.3 & 70.1 & 49.8 & -
        \\
        Grok & 44.5 & 44.5 & 42.3 & 45.4 & 45.7 & 76.5 & 49.5 & -
        \\
        Gemini & \textbf{50.2} & \ul{48.3} & \textbf{52.8} & \textbf{48.5} & \textbf{51.0} & \textbf{82.6} & \ul{58.3} & -
        \\
        \midrule
        \multicolumn{9}{c}{\textit{Open-source DR Agents}}
        \\
        Tongyi & 41.8 & 42.2 & 39.2 & 43.6 & 42.2 & 58.6 & 41.2 & -
        \\
        Enterprise & 46.0 & 46.9 & 43.5 & 46.8 & 46.8 & 72.3 & 51.8 & -
        \\
        LangChain (Web) & 44.4 & 48.5 & 39.4 & 47.2 & 42.5 & 63.7 & 52.1 & -
        \\
        LangChain (Table) & 44.9 & 45.5 & 41.9 & 45.9 & 46.2 & 60.3 & 54.9 & 20.4
        \\
        LangChain (Hybrid) & 44.8 & 47.4 & 40.2 & 46.6 & 44.8 & 62.3 & 53.6 & \ul{21.2}
        \\
        ThinkDepth (Web) & 46.1 & 46.6 & 42.8 & 47.1 & 47.8 & 71.7 & 51.8 & -
        \\
        ThinkDepth (Table) & 46.1 & 46.5 & 43.2 & 47.0 & 47.6 & 67.9 & 53.2 & 20.1
        \\
        ThinkDepth (Hybrid) & 46.3 & 47.1 & 42.7 & 47.6 & 47.8 & 68.3 & 53.6 & 18.3
        \\
        \midrule
        \textbf{HKA} & \ul{48.4} & \textbf{48.6} & \ul{48.8} & \ul{47.7} & \ul{48.5} & \ul{82.1} & \textbf{61.7} & \textbf{27.8} \\
        \bottomrule
    \end{tabular}}
    \label{tab:main_result}
\end{table*}

\section{Experiments}
\label{sec:experiment}
We conduct experiments on KDR-Bench and compare HKA with several baseline methods, including LLMs with search tools and deep research agents, to address the following research questions:

\textbf{RQ1.} How does HKA compare with existing DR agents across the three metric categories?

\textbf{RQ2.} How do the key sub-agents and key steps contribute to HKA?

\textbf{RQ3.} How reliable is the proposed evaluation framework?

\textbf{RQ4.} How does HKA generate multimodal content step by step?

\subsection{Experimental Setup}

\subsubsection{Baselines}
We evaluate three categories of systems, including LLMs with search tools, closed-source DR agents, and open-source DR agents.
With respect to LLMs with search tools, we select Hunyuan2.0~\cite{tencent_hunyuan_20}, GLM4.6~\cite{5team2025glm45agenticreasoningcoding}, Qwen3-Max~\cite{qwen}, and Minimax-M2~\cite{minimax_m2}.
With respect to closed-source DR agents, we select OpenAI Deep Research (OpenAI)~\cite{openai_deep_research_system_card_2025}, Grok-4 DeeperSearch (Grok)~\cite{xai_grok4_model_card_2025}, Perplexity Deep Research (Perplexity)~\cite{perplexity_deep_research_blog_2025}, and Gemini-3-Pro Deep Research (Gemini)~\cite{google_gemini_deep_research_doc_2025}.
The above systems are configured to use their default search sources.
With respect to open-source DR agents, we select Tongyi-Deepresearch-30B-A3B (Tongyi)~\cite{tongyidr}, Enterprise Deep Research (Enterprise)~\cite{prabhakar2025enterprisedeepresearch}, LangChain Open Deep Research (LangChain)~\cite{langchain_open_deep_research_2025}, and ThinkDepthAI Deep Research (ThinkDepth)~\cite{thinkdepthai2025deep_research}.
Further, we modify LangChain and ThinkDepth to support three search settings: (1) Web Search, which only searches for web pages; (2) Table Search, which only searches for tables in KDR-Bench; and (3) Hybrid Search, which searches for both.
In the latter two settings, tables are treated as text and integrated into prompts, similar to web pages.

\subsubsection{Implementation Details}
HKA is implemented via the LangGraph framework~\footnote{https://github.com/langchain-ai/langgraph}.
We use Qwen3-235B-A22B-Instruct-2507~\cite{qwen3} as the backbone LLM of the Planner and the Writer.
For Structured Knowledge Analyzer, we use Qwen3-Coder-480B-A35B-Instruct~\cite{qwen3} as the code LLM, Qwen3-VL-235B-A22B-Instruct-2507~\cite{bai2025qwen3vltechnicalreport} as the vision-language model, text-embedding-3-small~\cite{OpenAI_text_embedding_3_2024} as the embedding model and FAISS~\cite{douze2024faiss} as the vector store.
For Unstructured Knowledge Analyzer, we use Serper~\cite{serperdev} as the web search tool, Crawl4AI~\cite{crawl4ai2024} as the crawler tool, and Qwen3-235B-A22B-Instruct-2507 as the backbone LLM.
For a fair comparison, the open-source DR agents (except Tongyi) are also configured with the same backbone LLM and the same search tool.

With respect to the KDR-Bench, we use Gemini-2.5-Flash~\cite{gemini2flash} in the Data Collection step to generate table summaries.
In the following two steps, we use Gemini-3-Pro~\cite{google_gemini_deep_research_doc_2025} to generate questions and annotate knowledge points.
To avoid potential bias, for general-purpose and knowledge-centric metrics, we use DeepSeek-V3.2~\cite{deepseekai2024deepseekv3technicalreport} as the judge LLM; for vision-enhanced metrics, we use GPT-5~\cite{openai2026gpt5} as the judge MLLM.

\subsection{Performance Comparison (RQ1)}
We compare HKA with aforementioned baselines across three metric categories.
For general-purpose and knowledge-centric metrics, we evaluate all baselines.
For vision-enhanced metrics, due to space limitations, we select 9 baselines.
The results are shown in Table~\ref{tab:main_result} and Table~\ref{tab:vision_result}.

\subsubsection{Results on General-Purpose Metrics}
The results on general-purpose metrics are shown in the left part of Table~\ref{tab:main_result}.
From these results, we have the following observations:

(1) HKA outperforms the baselines except Gemini, showing its effectiveness in utilizing both structured and unstructured knowledge to generate high-quality reports.
Notably, HKA outperforms LangChain and ThinkDepth by more than 2.1 points in average score, despite utilizing the same backbone LLM, which highlights the superiority of HKA.
Although HKA outperforms most closed-source DR agents, it still falls short of the best-performing DR agent, Gemini.
We suppose that the performance gap is largely due to the differences in the backbone LLM capabilities.

(2) Generally, Deep Research agents outperform LLMs with search tools, suggesting that multi-step reasoning enables more effective information integration and supports higher-quality report generation. 
Another interesting finding is that open-source DR agents show comparable performance to closed-source ones, except Gemini.
However, HKA outperforms both open-source and closed-source DR agents (except Gemini) significantly, which implies that enhancing the structured knowledge analysis capability is an effective way to break through the current bottleneck in DR, besides improving the backbone LLMs.

(3) DR agents with table search usually obtain higher Depth scores than those with web search, supporting our hypothesis that structured knowledge facilitates deeper analysis.
However, agents with hybrid search do not show significant improvements over those using web search or table search, suggesting that a shallow integration of unstructured and structured knowledge is insufficient.
In contrast, HKA more effectively utilizes both kinds of knowledge and achieves a higher average score.

\begin{table*}[h]
    \centering
    \caption{Results on Vision-Enhanced metrics. Greener cells indicate a larger advantage for HKA, while redder cells indicate a larger disadvantage for HKA.}
    \begin{tabular}{lcccccccccc}
        \toprule
        Methods & Agr. & P\&E & E\&E & F\&I & M\&E & Soc. & Art & Tech. & Trans. & Avg.
        \\
        \midrule
        GLM4.6 &
\cellcolor{green!45}91.7 &
\cellcolor{green!45}100.0 &
\cellcolor{green!45}100.0 &
\cellcolor{green!45}100.0 &
\cellcolor{green!45}100.0 &
\cellcolor{green!45}100.0 &
\cellcolor{green!30}83.3 &
\cellcolor{green!45}100.0 &
\cellcolor{green!45}100.0 &
\cellcolor{green!45}97.6 \\

Minimax-M2 &
\cellcolor{green!45}100.0 &
\cellcolor{green!30}80.0 &
\cellcolor{green!45}100.0 &
\cellcolor{green!30}83.3 &
\cellcolor{green!45}100.0 &
\cellcolor{green!30}83.3 &
\cellcolor{green!45}100.0 &
\cellcolor{green!45}100.0 &
\cellcolor{green!45}100.0 &
\cellcolor{green!45}92.2 \\

Perplexity &
\cellcolor{green!15}66.7 &
\cellcolor{green!45}91.7 &
\cellcolor{green!15}60.0 &
\cellcolor{green!15}66.7 &
\cellcolor{green!30}70.0 &
\cellcolor{green!30}83.3 &
\cellcolor{green!15}66.7 &
\cellcolor{gray!15}50.0 &
\cellcolor{green!45}100.0 &
\cellcolor{green!30}72.0 \\

Grok &
\cellcolor{green!15}66.7 &
\cellcolor{green!45}100.0 &
\cellcolor{green!45}100.0 &
\cellcolor{green!15}58.3 &
\cellcolor{green!30}80.0 &
\cellcolor{green!30}83.3 &
\cellcolor{green!15}66.7 &
\cellcolor{green!45}100.0 &
\cellcolor{green!45}100.0 &
\cellcolor{green!30}81.7 \\

Gemini &
\cellcolor{gray!15}50.0 &
\cellcolor{green!15}66.6 &
\cellcolor{green!15}60.0 &
\cellcolor{red!25}41.7 &
\cellcolor{green!15}60.0 &
\cellcolor{gray!15}50.0 &
\cellcolor{green!15}66.7 &
\cellcolor{gray!15}50.0 &
\cellcolor{green!45}100.0 &
\cellcolor{green!15}56.1 \\

Enterprise &
\cellcolor{green!15}66.7 &
\cellcolor{green!45}100.0 &
\cellcolor{green!45}90.0 &
\cellcolor{green!15}66.7 &
\cellcolor{green!45}100.0 &
\cellcolor{green!30}83.3 &
\cellcolor{green!30}83.3 &
\cellcolor{green!15}66.7 &
\cellcolor{green!45}100.0 &
\cellcolor{green!30}82.9 \\

LangChain (Web) &
\cellcolor{green!30}75.0 &
\cellcolor{green!45}100.0 &
\cellcolor{green!30}80.0 &
\cellcolor{green!30}75.0 &
\cellcolor{green!45}100.0 &
\cellcolor{green!30}83.3 &
\cellcolor{green!45}100.0 &
\cellcolor{gray!15}50.0 &
\cellcolor{green!45}100.0 &
\cellcolor{green!30}84.1 \\

LangChain (Table) &
\cellcolor{green!30}83.3 &
\cellcolor{green!15}60.0 &
\cellcolor{green!45}100.0 &
\cellcolor{green!15}66.7 &
\cellcolor{green!30}80.0 &
\cellcolor{green!30}83.3 &
\cellcolor{green!15}66.7 &
\cellcolor{green!15}66.7 &
\cellcolor{green!45}100.0 &
\cellcolor{green!30}77.5 \\

LangChain (Hybrid) &
\cellcolor{green!30}75.0 &
\cellcolor{gray!15}50.0 &
\cellcolor{green!30}80.0 &
\cellcolor{gray!15}50.0 &
\cellcolor{green!45}100.0 &
\cellcolor{green!45}100.0 &
\cellcolor{green!45}100.0 &
\cellcolor{green!15}66.7 &
\cellcolor{green!45}100.0 &
\cellcolor{green!30}76.8 \\
        
        \bottomrule
    \end{tabular}
    \label{tab:vision_result}
\end{table*}

\subsubsection{Results on Knowledge-Centric Metrics}
The results on knowledge-centric metrics are shown in the right part of Table~\ref{tab:main_result}.
From these results, we have the following observations:

(1) HKA outperforms the baselines, except Gemini, by more than 5.6 points in Main Conclusion Alignment, demonstrating its effectiveness in producing high-level conclusions for report generation.
In contrast, although HKA underperforms Gemini, the gap is marginal, indicating that HKA achieves similar performance to the best-performing DR agent.
This is because the main conclusion integrates unstructured and structured knowledge, as mentioned in Section~\ref{sec:knowledge_centric_metrics}.
Gemini can obtain a high Main Conclusion Alignment score even though it relies primarily on unstructured knowledge sources.
Another supporting evidence is that LangChain (Table) and ThinkDepth (Table) exhibit lower Main Conclusion Alignment scores than LangChain (Web) and ThinkDepth (Web), respectively.

(2) HKA outperforms all the baselines, including Gemini, by more than 3.4 points in Key Point Coverage, demonstrating its effectiveness in analyzing structured knowledge.
As described in Section~\ref{sec:knowledge_centric_metrics}, Key Point Coverage emphasizes measuring the structured knowledge utilization of DR agents.
Therefore, HKA obtains the best Key Point Coverage score via using code to analyze structured knowledge in-depth and using a vision-language model to produce insights about it.
Similarly, LangChain and ThinkDepth with table and hybrid search, both of which utilize structured knowledge, also demonstrate higher scores than their web search counterparts.

(3) HKA outperforms all the baselines by more than 6.6 points in Key Point Supportiveness.
This is because HKA separately analyzes structured and unstructured knowledge, and then composes the results via the Writer, which reduces mutual interference between the two knowledge sources.
In contrast, agents with hybrid search (i.e., shallow integration of structured and unstructured knowledge) do not yield significant gains over their table-search counterparts and may even hinder structured knowledge utilization in ThinkDepth.

\subsubsection{Results on Vision-Enhanced Metrics}
As shown in Table~\ref{tab:vision_result}, we calculate pairwise win rates between HKA and other baselines.
In general, HKA outperforms all the baselines, including Gemini, in terms of win rates.
The inconsistency between General-Purpose (50.2 for Gemini and 48.4 for HKA) and vision-enhanced results (56.1 for HKA vs. Gemini) highlights the importance of visual features, such as the layout and figure content.
Notably, HKA not only generates figures from tables, but also cites external figure links extracted from web pages.
On average, HKA generates 5.75 figures from tables and cites 0.98 figures from web pages.
This suggests that the Structured Knowledge Analyzer significantly improves the visual presentation of the generated reports.

On average, HKA shows the largest advantage in the Trans. domain and the least advantage in the F\&I domain.
To further investigate the reason for this phenomenon, we count the number of figures and tables in the generated reports.
Results show that HKA generates 9.50 figures and tables per Trans. report, whereas Gemini only generates 2.00 per report.
In contrast, HKA generates 9.67 figures and tables per F\&I report, and Gemini generates 5.17 per report.
We hypothesize this is because, in the F\&I domain, web pages contain rich structured knowledge such as tables, which allows baselines to achieve better visual presentation by generating more tables.
This limits the advantage HKA gains from its figure generation capability.

\subsection{Ablation Study (RQ2)}

\begin{table}[]
    \centering
    \caption{Results of ablation study.}
    \begin{tabular}{lccc}
        \toprule
        & G.P. Avg. & Main. & Key.
        \\
        \midrule
        HKA & 48.4 & 82.1 & 61.7 \\
        wo. S.K.A. & 46.7$^{\downarrow 1.7}$ & 74.1$^{\downarrow 8.0}$ & 52.7$^{\downarrow 9.0}$ \\
        wo. U.K.A. & 45.4$^{\downarrow 3.0}$ & 75.2$^{\downarrow 6.9}$ & 58.7$^{\downarrow 3.0}$ \\
        wo. Rerank & 46.1$^{\downarrow 2.3}$ & 79.3$^{\downarrow 2.8}$ & 59.9$^{\downarrow 1.9}$ \\
        wo. Schema & 48.1$^{\downarrow 0.3}$ & 81.7$^{\downarrow 0.4}$ & 60.3$^{\downarrow 1.4}$ \\
        \bottomrule
    \end{tabular}
    \label{tab:ablation_result}
\end{table}

To investigate the effectiveness of key sub-agents and key steps, we conduct an ablation study on HKA.
We investigate four variants of HKA: (1) removing Unstructured Knowledge Analyzer, which is denoted as \textit{wo. U.K.A.}; (2) removing Structured Knowledge Analyzer, which is denoted as \textit{wo. S.K.A.}; (3) removing the rerank step in knowledge retrieval and directly obtaining the most relevant table via vector similarity, which is denoted as \textit{wo. Rerank}; and (4) replacing the comment-style schema with comment-style table data, which is denoted as \textit{wo. Schema}.
We report three representative metrics, i.e., average score on general-purpose metrics (denoted as \textit{G.P. Avg.}), Main Conclusion Alignment, and Key Point Coverage.

As shown in Table~\ref{tab:ablation_result}, when we remove either Unstructured Knowledge Analyzer or Structured Knowledge Analyzer, performance in all three metrics declines significantly.
These results imply that both kinds of knowledge are essential for HKA to produce high-quality reports.
When we remove the rerank step in knowledge retrieval, HKA also exhibits a performance drop, which implies that vector-based retrieval alone seems insufficient for selecting appropriate tables for different subtasks.

Compared to above variants, replacing the comment-style schema with table data leads to only a slight performance drop.
These results suggest that the code LLM is able to organize raw data and perform appropriate computation over them.
Notably, our schema-based approach becomes a more practical choice as the scale of table data increases, since the schema provides a precise and compact description of the data fields for operating on structured knowledge.

\subsection{Evaluation Reliability Study (RQ3)}

We analyze the reliability of the proposed evaluation framework from three aspects, namely, the judge model, the score consistency, and the human preference.
We choose three representative metrics, namely, average score of General-Purpose metrics (G.P. Avg.), Main Conclusion Alignment (Main.), and Key Point Coverage (Key.).

\subsubsection{Judge Model Analysis}

\begin{table}[h]
    \centering
    \caption{Results on different judge LLMs.}
    \begin{tabular}{lccc}
        \toprule
        Method & Metric & Default & Claude 
        \\
        \midrule
        \multirow{3}{*}{Minimax-M2} & G.P. Avg. & 44.7 & 45.6 
        \\
        & Main. & 58.2 & 58.1 
        \\
        & Key. & 50.5 & 50.5 
        \\
        \midrule
        \multirow{3}{*}{Perplexity} & G.P. Avg. & 46.5 & 46.8 
        \\
        & Main. & 70.1 & 67.1 
        \\
        & Key. & 49.8 & 50.9 
        \\
        \midrule
        \multirow{3}{*}{Enterprise} & G.P. Avg. & 46.0 & 46.3 
        \\
        & Main. & 72.3 & 70.1 
        \\
        & Key. & 48.4 & 47.9 
        \\
        \midrule
        \multirow{3}{*}{HKA} & G.P. Avg. & 48.4 & 48.6 
        \\
        & Main. & 82.0 & 78.1 
        \\
        & Key.& 61.7 & 61.3 
        \\
        \bottomrule 
    \end{tabular}
    \label{tab:judge_model_result}
\end{table}

To examine whether different judge LLMs affect the evaluation results, we replace the default judge LLM (Default) with Claude-haiku-4.5-thinking (Claude)~\cite{anthropic_claude_haiku_45_thinking}.
As shown in Table~\ref{tab:judge_model_result}, the two judge LLMs derive relatively consistent rankings of the DR agents on G.P. Avg. and Main.
With respect to Key., although the ranking of Minimax-M2 and Perplexity is opposite between the two judges, the absolute scores are similar, which demonstrates the stability of the evaluation.

\subsubsection{Score Consistency Analysis}

\begin{table}[h]
    \centering
    \caption{Results on different generation runs.}
    \begin{tabular}{lccccc}
        \toprule
        Method & Metric & Run 1 & Run 2 & Run 3 & S.D.
        \\
        \midrule
        \multirow{3}{*}{Minimax-M2} & G.P. Avg. & 44.7 & 45.1 & 44.6 & 0.2
        \\
        & Main. & 58.2 & 58.9 & 58.7 & 0.3
        \\
        & Key. & 50.5 & 49.7 & 50.1 & 0.3
        \\
        \midrule
        \multirow{3}{*}{Enterprise} & G.P. Avg. & 46.0 & 46.2 & 45.6 & 0.2
        \\
        & Main. & 72.3 & 73.1 & 71.6 & 0.6
        \\
        & Key. & 51.8 & 50.6 & 51.8 & 0.6
        \\
        \midrule
        \multirow{3}{*}{HKA} & G.P. Avg. & 48.4 & 48.3 & 48.0 & 0.2
        \\
        & Main. & 82.0 & 82.1 & 81.6 & 0.2
        \\
        & Key. & 61.7 & 63.0 & 62.5 & 0.5
        \\
        \bottomrule 
    \end{tabular}
    \label{tab:score_consistency_result}
\end{table}

To examine whether scores vary significantly across different generation runs, we generate reports in three independent runs and calculate the standard deviation (S.D.).
Considering the time and financial costs, we exclude closed-source DR agents from this analysis.
As shown in Table~\ref{tab:score_consistency_result}, among the three metrics, G.P. Avg. exhibits the highest stability (0.2 for Minimax-M2, Enterprise, and HKA).
In contrast, Key. shows the largest S.D. (0.3 for Minimax-M2, 0.6 for Enterprise, and 0.5 for HKA), probably because it relates to fine-grained knowledge (tables) and is more easily affected by the randomness of the agents' actions.
Nevertheless, the overall S.D. remains low ($\le 0.6$), suggesting that these metrics are mostly consistent across runs.

\subsubsection{Human Preference Analysis}
To examine whether the proposed evaluation framework aligns with human preferences, we randomly select 10 reports from HKA, Gemini, Grok, Perplexity, LangChain (Web), and LangChain (Table).
Following the same evaluation guidelines used by the MLLM judge, a human judge is asked to determine the relative quality of each given pair of reports.
We pair HKA and Gemini with each of the other four agents, resulting in 80 report pairs for comparison.
The pairwise agreement rate (PAR) is calculated between the human preferences and the vision-enhanced results for these report pairs.
The result is 86.3\%, which shows a relatively high consistency between the proposed evaluation framework and human preferences.

\subsection{Case Study (RQ4)}

\begin{figure}
    \centering
    \includegraphics[width=\linewidth]{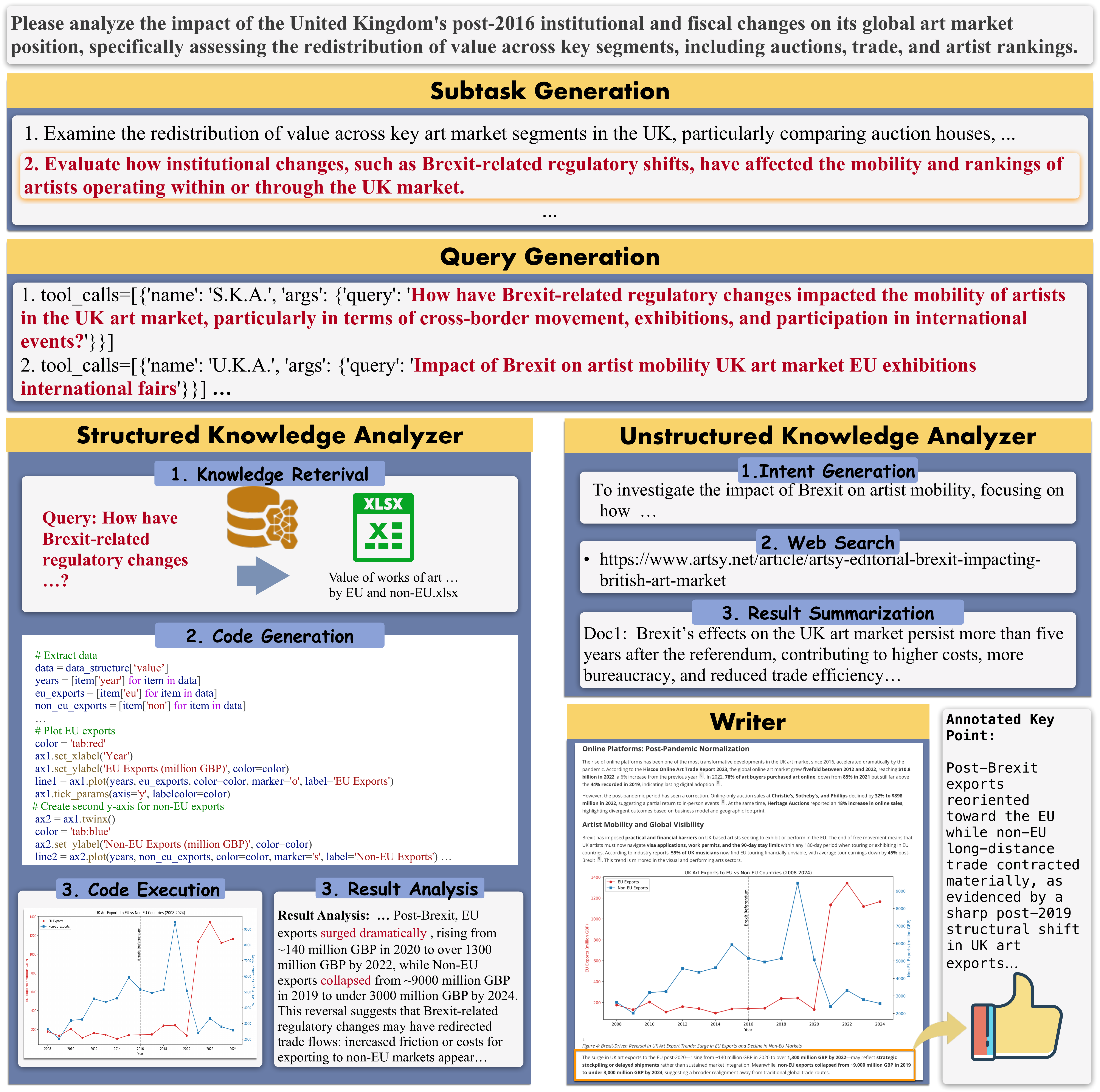}
    \caption{Case study for HKA.}
    \label{fig:case_study}
\end{figure}

To further investigate the details of HKA, we conduct a case study on it.
Figure~\ref{fig:case_study} demonstrates part of the research trajectory for a question about Art.
As shown in the figure, the Planner first decomposes the question into several subtasks.
In the subtask ``Evaluate how institutional changes, such as Brexit-related regulatory shifts ...'', it invokes the Structured and Unstructured Knowledge Analyzers via two tool calls, respectively.
The Structured Knowledge Analyzer then retrieves the table about ``How have Brexit-related regulatory changes impacted the mobility of artists in the UK art market...'' and generates a figure and the corresponding insights about the differences between developed and developing countries. And Unstructured Knowledge Analyzer further search the ``Impact of Brexit on artist mobility UK art market EU exhibitions international fairs''.
The Writer inserts the figure into the final report and integrates the corresponding insights into the textual content.
Finally, the judge LLM recognizes that this text snippet matches an annotated key point. This case suggests the effectiveness of HKA and the evaluation framework.

\section{Conclusion}\label{sec:conclusion} 
In this paper, we introduced the Knowledgeable Deep Research (KDR) task, which requires deep research agents to generate reports grounded in both structured and unstructured knowledge. 
To tackle this task, we proposed the Hybrid Knowledge Analysis framework (HKA), a multi-agent architecture that reasons over both types of knowledge and composes generated text, figures, and tables into multimodal reports.
To support evaluation, we constructed KDR-Bench, which spanned 9 domains and comprised 41 expert-level questions with 1,252 reference tables, together with an LLM-based evaluation framework.
Experimental results demonstrated that KDR remained challenging for existing DR agents, while HKA consistently outperformed strong baselines, highlighting the importance of explicit, data-driven analysis for future DR systems.

\bibliographystyle{ACM-Reference-Format}
\bibliography{sample-sigconf}


\end{document}